\RequirePackage{fix-cm}
\documentclass[smallextended]{svjour3}       % onecolumn (second format)
\usepackage{graphicx}
\usepackage{epsfig}
\usepackage{natbib}
\usepackage[utf8]{inputenc}
\usepackage{amsmath, amssymb}
\usepackage{graphicx}
\usepackage{fancyhdr}
\usepackage[pdftitle={Federico Schl\"uter, thesis proposal}]{hyperref} 
\usepackage{geometry}
\usepackage[T1]{fontenc}
\usepackage{color}
\usepackage[ruled]{algorithm}
\usepackage{algorithmic}
\usepackage{makeidx}
\usepackage{multirow}

\newtheorem{openproblem}{Open Problem}
\hypersetup{linktocpage}

\newcommand{\set}[1]{\ensuremath{\mathbf{#1}}}

\newcommand{\PERP}{\mbox{\ensuremath{\perp\!\!\!\perp}}} 
\newcommand{\NPERP}{\mbox{\ensuremath{\,\not\!\perp\!\!\!\perp}}}

\newcommand{\indep}[2]{\ensuremath{#1 \PERP #2}}
\newcommand{\dep}[2]{\ensuremath{#1 \NPERP #2}}
\newcommand{\ci}[3]{\ensuremath{(\indep{#1}{#2} | #3})}
\newcommand{\cd}[3]{\ensuremath{(\dep{#1}{#2} | #3})}
\newcommand{\triplet}[3]{\ensuremath{\langle{#1},{#2}, #3} \rangle}
\newcommand{\blanket}[1]{\ensuremath{\set{MB}^{#1}}}

\newcommand{\data}{\ensuremath{D}}

\newcommand{\parameters}{\ensuremath{\boldsymbol{\theta}}}

\begin{document}

\title{A survey on independence-based Markov networks learning}
\author{Federico Schl\"uter}

\institute{F. Schl\"uter \at
              \email{federico.schluter@frm.utn.edu.ar}  \\
	  Home page: {http://dharma.frm.utn.edu.ar/fschluter/} \\
	  Lab. DHARMa of Artificial Intelligence, \\
	  Dept of Information Systems, Facultad Regional Mendoza, National Technological University, Argentina. 
}

\date{Received: date / Accepted: date}

\maketitle

\begin{abstract}
The problem of \emph{learning} the \emph{Markov network structure} from data
has become increasingly important in machine learning,
and in many other application fields.
Markov networks are \emph{probabilistic graphical models},
a widely used formalism for handling probability distributions in intelligent systems.
This document focuses on a technology called \emph{independence-based} learning,
which allows for the learning of the independence structure of Markov networks from data in an efficient and sound manner, 
whenever the dataset is sufficiently large, and data is a representative sample of the target distribution.
In the analysis of such technology, this work surveys the current state-of-the-art algorithms,
discussing its limitations, and posing a series of open problems 
where future work may produce some advances in the area, in terms of quality and efficiency.
\end{abstract}

\section{Motivation \label{sec:introduction}}
Nowadays intelligent systems have to reason in realistic domains, 
storing knowledge of the world, and supporting efficient inference, even when exceptions occur.
This is referred to in the literature as \emph{reasoning under uncertainty}.
A popular approach taken for reasoning under uncertainty is the use of 
\emph{probabilistic models}, a statistical analysis tool 
for \emph{statistical inference}.
The statistical inference process is used for drawing conclusions from data 
by calculating the probability of propositional sentences.
An example of a probabilistic model is the \emph{tabular probabilistic model},  
a function represented as a table that
assigns a probability to every possible complete assignment in a domain, 
so that the sum of the probabilities adds up to 1.
Figure~\ref{fig:tabularmodel} illustrates an abstract tabular model 
for a domain with $n$ binary variables $\set{V}=\{X_0,...,X_{n-1}\}$,
consisting on $2^n$ tuples, one per possible configuration of variables. 

%.....................................................
\begin{figure}[h]
\begin{center}
	\begin{tabular}{cccccc}
	\hline
	\textbf{$X_0$ } &  \textbf{$X_1$}  & ... & \textbf{$X_{n-1}$}  & & \textbf{ $ \Pr(X_0,..,X_{n-1})$}  \\
	\hline
	0 & 0 & ... & 0 & & 0.121 \\
	0 & 0 & ... & 1 & & 0.076 \\
	. & . & ... & . & & . \\
	. & . & ... & . & & . \\
	. & . & ... & . & & . \\
	1 & 1 & ... & 0 & & 0.21 \\
	1 & 1 & ... & 1 & & 0.12 \\
	\hline
	\end{tabular}

\end{center} 

\caption{\label{fig:tabularmodel}An example tabular model over $n$ binary random variables, with $2^n$ numerical parameters.}

\end{figure}

However, the tabular model presents computational and semantic limitations.
First, its storage requirements are exponential in the number of variables, 
and the size of its respective domains.
When domains of variables are continuous, such tables are infinite,
and in practice some mathematical functions can be used.
Nonetheless, in this work the attention is restricted only to discrete distributions, 
so continuous variables may be considered as discrete variables.
Second, queries of interest usually do not involve all the variables,
and the cost of computing marginal and conditional probabilities would result 
in exponential summations of variable combinations. 
Third, such representation has no clear semantics for humans. 
The common pattern of human knowledge has
probabilistic judgments for a small number of propositions.
Therefore, \emph{conditional independences} are a natural 
way of representing probability distributions.
It is common for people to judge a three-place relationship 
of conditional dependency, i.e., $X$ influences $Y$, given $Z$.

Using independences may reduce the exponential requirements of the tabular model. 
For example, making the simple assumption that all the $n$ variables in Figure~\ref{fig:tabularmodel} 
are \emph{mutually independent} allows decomposing the joint probability distribution as
\begin{equation*}
 \Pr(X_0,...,X_{n-1}) = \prod\limits^{n-1}_{i=0} {\Pr(X_i)}.
\end{equation*}
Such decomposition requires a polynomial number ($n$) of exponentially smaller tables with only two rows.
Figure~\ref{fig:mutuallyindependent} illustrates a model assuming that all the 
binary variables are mutually independent,
consisting only of $n$ tables with $2$ tuples each.

%.....................................................
\begin{figure}[ht]
\begin{center}
\vspace{0.5cm}
\begin{tabular}{c}
	\begin{tabular}{cc}
		\hline
	\textbf{$X_0$ } &  \textbf{ $\Pr(X_0)$}  \\
		\hline
	0 & 0.21 \\
	1 & 0.79 \\
		\hline
	\end{tabular} ~~~\textbf{,}~~~ 
	\begin{tabular}{cc}
		\hline
	\textbf{$X_1$} &  \textbf{$ \Pr(X_1)$}  \\
		\hline
	0 & 0.45 \\
	1 & 0.55 \\
		\hline
	\end{tabular}
	~~~\textbf{...}~~~  
	\begin{tabular}{cc}
		\hline
	\textbf{$X_{n-1}$} & \textbf{ $ \Pr(X_{n-1})$}  \\
		\hline
	0 & 0.42 \\
	1 & 0.58 \\
		\hline
	\end{tabular}

\end{tabular}
\end{center} 

\caption{\label{fig:mutuallyindependent}An example model assuming that all 
the variables of the domain are mutually independent, with $n$ tables of only $2$ numerical parameters each.}
\vspace{0.5cm}
\end{figure}

To address all these problems, namely the exponential storage requirements, 
the exponential cost of computing marginal and conditional probabilities, 
and the lack of explicitness of the model, 
several researchers in the late 80's created \emph{probabilistic graphical models}, or simply, 
\emph{graphical models}, a well-established formalism for representing compactly 
joint probability distributions. 
They are composed of an independence structure, 
and a set of numerical parameters.
The structure encodes the independences present in the distribution,
and then defines a family of probability distributions.
The set of numerical parameters defines a
unique distribution among this family and 
quantifies the relationships in the structure.
Such representation is explained in more detail in Section~\ref{sec:markovnetworks}.

The most important types of graphical models are \emph{Bayesian networks} 
and \emph{Markov networks} \cite{pearl88}.
% , and more recently \emph{Factor Graphs} \cite{Kschischang01,Abbeel06}.
The well-known Bayesian networks are graphical models for encoding distributions 
where dependencies are representable by a directed acyclic graph. 
Markov networks (also known as \emph{Markov Random Fields}, \emph{undirected graphical models}, 
or simply \emph{undirected models}) encode distributions 
where dependencies are representable by an undirected graph.
Three of the most influential textbooks on this topic published 
in the last three decades are \cite{pearl88}, 
\cite{LAURITZEN96} and \cite{koller09}.

There have been many applications of graphical models 
in a wide range of fields during recent years.
Some examples are present in the areas of computer vision and image analysis.
\cite{BESAG91} gives two examples, one in archeology, the other in epidemiology;
in \cite{ANGUELOVTASKAR05}, addressing the problem of segmenting 3D scan data into objects or object classes;
or \cite{Li2006}, a complete textbook that presents an exposition of Markov Random fields 
to image restoration and edge detection in the low-level domain, 
and object matching and recognition in the high-level domain. 
More examples are present in the area of spatial data mining and geostatistics,
as those presented in the textbook of ~\cite{cressie92}, where 
Markov Random Fields are emphasized for modeling spatial lattice data; 
or more recently, the work of ~\cite{shekhar04} 
that presents spatial analysis methods and applications for Markov Random Fields in a wide range of fields,
including biology, spatial economics, environmental and earth science, ecology, 
geography, epidemiology, agronomy, forestry and mineral prospection.
There are also examples for disease diagnosis, such as \cite{schmidts08} 
that presents a Markov Random Fields based method for detecting coronary heart disease
processing ultrasound images of echocardiograms. Also  
in the area of computational biology, \cite{friedman00}
proposes the use of Bayesian networks for discovering interactions among genes.
More applications of graphical models are present for 
evolutive optimization searching,
as \cite{larranagaANDlozano2002} that describes the use of Bayesian networks for modeling
the probability distribution of individuals with high fitness in evolutive algorithms,
or more recently, \cite{Alden07} and \cite{moapaper} proposing Markov networks for the same purpose.
Further examples are shown for Information Retrieval by \cite{Metzler05amarkov} and \cite{cai07}, 
to model term dependencies using Markov Random Fields; 
and for malware propagation, \cite{Karyotis2010}
analyzes the spatial and contextual dependencies of malware propagation, 
also using Markov Random Fields.
There are many other interesting examples that could be included in this list.
Table~\ref{table:apps} summarizes all these examples
in order to help the readers to choose which method can be a better solution for a certain application.

\begin{table}[ht]
\scriptsize
\center
	      \caption{Some example Probabilistic Graphical Models applications. \label{table:apps}}
		      \begin{tabular}{p{11cm}  p{3cm}}
			    \multicolumn{2}{c}{~}\\
			    \hline
			    \\
			    \textbf{Application} & \textbf{Reference} \\ 
			    \\ \hline
			    \\

			    \begin{minipage}{11cm}
					\begin{itemize}% \setlength{\itemsep}{-2mm} 
					\item Computer vision, and image analysis, with examples in archeology and epidemiology
					\end{itemize}
					\end{minipage} & \cite{BESAG91} \\ 
			    \\ \hline
			    \\   
			    \begin{minipage}{11cm}
					\begin{itemize} % \setlength{\itemsep}{-2mm} 
					\item Segmentition of 3D scan data into objects 
					\end{itemize}
					\end{minipage} & \cite{ANGUELOVTASKAR05} \\ 
			    \\ \hline
			    \\
			    \begin{minipage}{11cm}
					\begin{itemize} % \setlength{\itemsep}{-2mm} 
					\item Markov Random Fields for image restoration, edge detection and object matching
					\end{itemize}
					\end{minipage} & \cite{Li2006} \\ 
			    \\ \hline
			    \\
			    \begin{minipage}{11cm}
					\begin{itemize} % \setlength{\itemsep}{-2mm} 
					\item Markov Random Fields for spatial data minging and geostatistics
					\end{itemize}
					\end{minipage} & \cite{cressie92} \\ 
			    \\ \hline
			    \\
			    \begin{minipage}{11cm}
					\begin{itemize} % \setlength{\itemsep}{-2mm} 
					\item Markov Random Fields for spatial spatial analysis methods in 
					biology, spatial economics, environmental and earth science, ecology, 
					geography, epidemiology, agronomy, forestry and mineral prospection.
					\end{itemize}
					\end{minipage} & \cite{shekhar04}\\ 

			    \\ \hline
			    \\
			    \begin{minipage}{11cm}
					\begin{itemize} % \setlength{\itemsep}{-2mm} 
					\item Markov Random Fields method for detecting coronary heart disease.
					\end{itemize}
					\end{minipage} & \cite{schmidts08} \\ 
			    \\ \hline
			    \\
			    \begin{minipage}{11cm}
					\begin{itemize} % \setlength{\itemsep}{-2mm} 
					\item Computational biology. Learning of Bayesian networks \\
					for discovering interactions among genes.
					\end{itemize}
					\end{minipage} & \cite{friedman00} \\ 
			    \\ \hline
			    \\
			    \begin{minipage}{11cm}
					\begin{itemize} % \setlength{\itemsep}{-2mm} 
					\item Bayesian networks for evolutive optimization search. 
					\end{itemize}
					\end{minipage} & \cite{larranagaANDlozano2002} \\ 
			    \\ \hline
			    \\
			    \begin{minipage}{11cm}
					\begin{itemize} % \setlength{\itemsep}{-2mm} 
					\item Markov networks for evolutive optimization search. 
					\end{itemize}
					\end{minipage} & \cite{Alden07,moapaper} \\ 

			    \\ \hline
			    \\
			    \begin{minipage}{11cm}
					\begin{itemize} % \setlength{\itemsep}{-2mm} 
					\item Information Retrieval. Markov Random Fields for modeling terms dependency.
					\end{itemize}
					\end{minipage} & \cite{Metzler05amarkov,cai07} \\ 

			    \\ \hline
			    \\
			    \begin{minipage}{11cm}
					\begin{itemize} % \setlength{\itemsep}{-2mm} 
					\item Malware propagation. Markov Random Fields for modeling spatial and contextual dependencies.
					\end{itemize}
					\end{minipage} & \cite{Karyotis2010} \\ 
			    \\ \hline
			    \\
			\end{tabular}

\end{table}

The framework provided by probabilistic graphical models supports
three critical capabilities of intelligent systems, 
as highlighted in the textbook of \cite{koller09}:  

\begin{itemize}

% % % % % % % % 	
  \item \textit{Representation}: a compact and declarative model of the
  knowledge based on graphs.
  On the one hand, such models are compact, providing a representation 
  of conditional independences present in a probability distribution 
  which is efficient and computationally tractable. 
  The compact representation of graphical models is achieved by exploiting a
  principle property present in many distributions: 
  variables tend to interact directly with very few others.
  On the other hand, since the models are graphical, they are declarative, 
  and a human expert can understand and evaluate their semantics and properties. 

% % % % % % % % % % % % 
  \item \textit{Inference}: given a graphical model, the most fundamental 
  and yet highly non-trivial task is to compute marginal distributions of one or a few variables. 
  This task is usually called \emph{inference}. 
  Through marginalization it is possible to compute conditionals and posteriors, and to make predictions.
  Inference is also a sub-routine of learning tasks, and is therefore the most 
  elementary sub-routine of graphical models.
  However, as proven by \cite{Cooper1990}, exact inference is NP-hard in general. 
  There are several methods for working directly with the structure of graphical models,
  that are in practice orders of magnitude faster 
  than manipulating explicitly the joint probability distribution.
  The textbook of \cite{koller09} provides an extensive discussion on this topic, and describes
  the most popular methods used, such as \emph{variable elimination},
  \emph{Monte Carlo methods}, and \emph{loopy belief propagation}.
  Other recent works are \emph{tree-reweighted message-passing} of \cite{Wainwright03tree-reweightedbelief}, 
  \emph{Power EP} of \cite{minka2004}, \emph{generalized belief
 propagation} of \cite{Yedidia04constructingfree}, and \emph{Variational message-passing} of \cite{Winn2005:VMP}.
  A free and open source library, providing implementations 
  of various exact and approximate inference methods for graphical models, was published recently by \cite{Mooij08libdai}.

% % % % % % % % % % 
  \item \textit{Learning}: constructing graphical models can be done
  either by a human expert or by learning it automatically from data.
  There are many algorithms that model the probability distribution of historical
  data, returning a graphical model as the solution.
  They are really useful, since experts knowledge is not always enough to design a proper
  model. Therefore, some authors consider these algorithms a tool for knowledge
  discovery. Moreover, when constructing models for a specific problem, it is possible to use
  the data-driven approach, using some part of the model provided by an expert 
  and filling the details automatically, by fitting the model to data. 
  The large number of success stories claimed using this approach in recent years has resulted in some 
  authors, such as Koller and Friedman, claiming that models produced by this process 
  are usually much better than those purely hand constructed. 

\end{itemize}

In this work, the specific problem of learning the independence structure of Markov networks is reviewed. 
This is an interesting problem that has resulted in important
contributions to this domain in recent years, although many of its core
challenges remain unresolved and are under intense deliberation.
This work focuses on a technology called independence-based learning,
which allows one to infer the independence structure of Markov networks from data in an efficient and sound manner, 
whenever data is sufficient and a representative sample of the target distribution.
An analysis of the current state-of-the-art algorithms
for learning Markov networks structure using such technology is presented,
discussing its current limitations, and its potential 
for improving the quality and the efficiency of current approaches.

The rest of the document is structured as follows: 
Section~\ref{sec:markovnetworks} presents an overview of \textit{Markov networks representation}. 
Section~\ref{sec:structurelearning} discusses the problem of \textit{learning Markov networks} from data. 
Section~\ref{sec:approaches} provides a review of current independence-based Markov network structure
learning algorithms. 
Finally, Section~\ref{sec:analysis} analyzes the surveyed independence-based algorithms
and discusses their relative advantage as well as disadvantages, 
concluding with a series of open problems that remain in the domain 
of independence-based structure learning for Markov networks.

% \newpage
\section{Markov networks representation\label{sec:markovnetworks}}
This section provides an overview of the representation of a specific type of graphical models: Markov networks.
Graphical models in general consist of a qualitative and a quantitative component
for representing a probability distribution $P$. 
Such distribution is given over a domain of $n$ variables, denoted $\set{V}=\{X_0,...,X_{n-1}\}$.
The qualitative component is the \emph{independence structure} $G$
(also known as the \textit{network}, or the \textit{graph}) of the model, 
that represents conditional independences among the domain variables,
and then defines a family of probability distributions.
The quantitative component is a set of numerical parameters $\parameters$, 
that defines a unique distribution among this family and 
quantifies the relationships in the structure.

\subsection{The independence structure \label{sec:independencestructure}}

The independence structure is a compact representation 
of \emph{conditional independences} present in the underlying distribution $P$. 
Two variables $X$ and $Y$ are independent conditioned in a set of variables $\set{Z}$
when knowing the value of $Y$ tells me nothing new about $X$, 
if I already know the values of variables in $\set{Z}$. 
In this work, conditional independence is denoted as $\ci{\set{X}}{\set{Y}}{\set{Z}}$,
and $\cd{\set{X}}{\set{Y}}{\set{Z}}$ denotes conditional dependence.

\begin{figure}[h]
	\begin{center}
		\includegraphics[height=6cm]{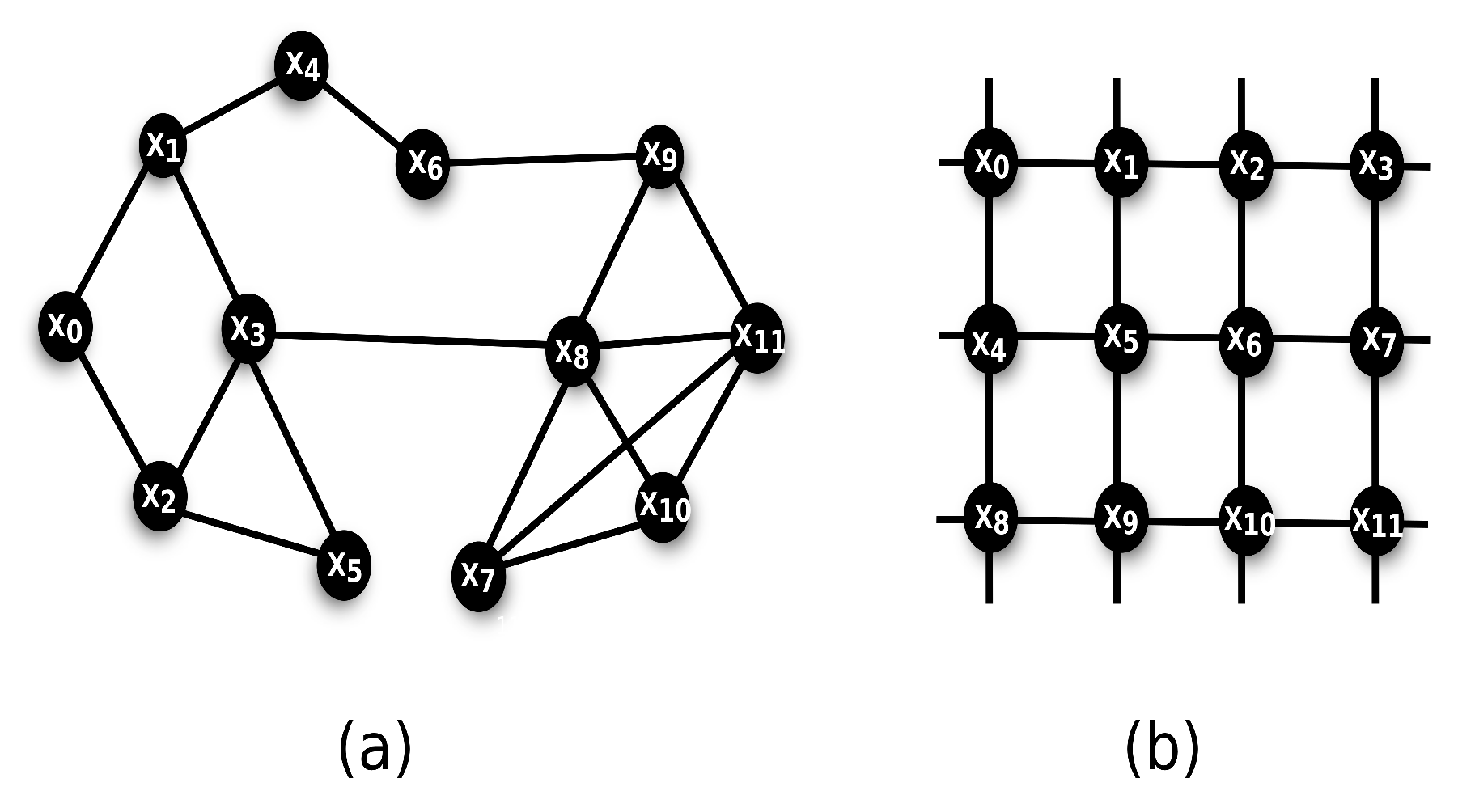}
	\end{center}
\caption{\label{fig:undirectedgraphs} Two examples of undirected independence structure:  
(a) an irregular graph with different grade of connectivity for distinct nodes, and  
(b) a regular lattice where variables belong to a domain in a spatial problem.}
\vspace{0.5cm}
\end{figure}
The structure $G$ of a Markov network is an undirected graph with $n$ nodes, 
each one representing a random variable in the domain.
The edges in the graph encode conditional independences among the variables.
Figure~\ref{fig:undirectedgraphs} shows two examples of undirected structures,
both representing domains with $n=12$, and variables $\set{V}=\{X_0, \dots , X_{11}\}$. 
The first example in Figure~\ref{fig:undirectedgraphs}~(a) is an irregular graph 
with different grade of connectivity for distinct nodes. 
The second in Figure~\ref{fig:undirectedgraphs}~(b) is a regular lattice 
where variables belong to a domain in a spatial problem, 
as typically used for representing 2D images, 
or for two dimensional Ising spin glasses models 
(mathematical models of ferromagnetism in statistical mechanics). 

The independence structure is a map of the independences in the underlying distribution, 
and such independences can be read from the graph through
\emph{vertex separation}, considering that each variable is conditionally independent 
of all its non-neighbor variables in the graph, 
given the set of its neighbor variables. This is called the Local Markov property.
For example, in Figure~\ref{fig:undirectedgraphs}~(a) variables $X_0$ and $X_3$ are conditionally independent, 
given the set of variables $\{X_1,X_2\}$. 
In the toroidal lattice of Figure~\ref{fig:undirectedgraphs}~(b), 
$X_5$ is conditionally independent of all the non-adjacent variables, 
given its neighbor variables $\{X_1,X_4,X_6,X_9\}$.

\subsubsection{Correctness of the structure \label{sec:correctness}}
% when the independences of p may be represented by G?
For correctly representing a probability distribution $P$ by a Markov network, 
$G$ must be a map of the independences present in $P$.
As proved in \cite{pearl88}, a graph $G$ is called an \emph{independence-map} (or \emph{I-map}, for short) 
of a distribution $P$ when all the independences encoded in the graph exist in the underlying distribution $P$.

\begin{definition}{I-map \cite{pearl88}}[p.92].
      
    A graph $G$ is an \emph{I-map} of a distribution $P$ if for all disjoint subsets of variables 
$\set{X}$, $\set{Y}$ and $\set{Z}$, the following is satisfied:
      \begin{equation}
      {\ci{\set{X}}{\set{Y}}{\set{Z}}}_{G} \Rightarrow {\triplet{\set{X}}{\set{Y}}{\set{Z}}}_{P},
      \end{equation}
    where ${\ci{\set{X}}{\set{Y}}{\set{Z}}}_{G}$ are the independences encoded by $G$,
    and ${\triplet{\set{X}}{\set{Y}}{\set{Z}}}_{P}$ are the independences existent in the underlying distribution $P$.

Similarly, $G$ is a dependency-map (D-map) when 
      \begin{equation}
      {\ci{\set{X}}{\set{Y}}{\set{Z}}}_{G} \Leftarrow {\triplet{\set{X}}{\set{Y}}{\set{Z}}}_{P}.
      \end{equation}

\end{definition}

Using a graph $G$ that is an I-map guarantees that nodes found to be separated correspond to independent variables,
but does not guarantee that all those showed to be connected are dependent.
Conversely, when $G$ is a D-map it is guaranteed that the nodes connected in $G$ are dependent in the distribution $P$.
Fully-connected graphs are trivial I-maps, and empty graphs are trivial D-maps.
A distribution $P$ is said to be a \emph{perfect-map} of $P$ if it is both an I-map and a D-map.

An axiomatic characterization of the family of relations 
that are isomorphic to vertex separation in graphs is given by the concept 
of graph-isomorphism. Basically, a distribution $P$ is a graph-isomorph 
when its independences among variables 
can be encoded by an undirected graph. 

\begin{definition}{Graph-isomorphism \cite{pearl88}}[p.93].

    A distribution is said to be a \textbf{graph-isomorph} if there 
exists an undirected graph $G$ that is a \textbf{perfect-map} of $P$, i.e., 
for every three disjoint subsets $\set{X}$, $\set{Y}$ and $\set{Z}$, we have
      \begin{equation}
      {\ci{\set{X}}{\set{Y}}{\set{Z}}}_{G} \Longleftrightarrow {\triplet{\set{X}}{\set{Y}}{\set{Z}}}_{P}.
      \end{equation}
\end{definition}

A necessary and sufficient condition for a distribution $P$ to be a graph-isomorph is that 
${\triplet{\set{X}}{\set{Y}}{\set{Z}}}_{P}$ satisfies the following axioms of independences,
introduced by \cite{PearlPaz1985:graphoids}. 
There is another set of axioms for learning Bayesian networks, 
but it is omitted here. 

  \begin{tabular}{l p{10cm} l}

  & & \\	
	\textbf{Symmetry} & $\ci{\set{X}}{\set{Y}}{\set{Z}} \Leftrightarrow \ci{\set{Y}}{\set{X}}{\set{Z}} $ &\\
  & &\\
	\textbf{Decomposition} & $\ci{\set{X}}{\set{Y}~\cup~\set{W}}{\set{Z}} \Rightarrow \ci{\set{X}}{\set{Y}}{\set{Z}} ~\&~\ci{\set{X}}{\set{W}}{\set{Z}}$ &\\
  & & \\	
	\textbf{Transitivity} & $\ci{\set{X}}{\set{Y}}{\set{Z}} \Rightarrow \ci{\set{X}}{\lambda}{\set{Z}} ~\mbox{or}~ \ci{\lambda}{\set{Y}}{\set{Z}}$
  &
  \raisebox{0.2cm}[0mm]{
    \begin{minipage}{0.2cm}
    \begin{equation}\label{eq:generalaxioms}\end{equation}\vspace{0.3cm}
   \end{minipage}
  }\\
	\textbf{Strong union} & $\ci{\set{X}}{\set{Y}}{\set{Z}} \Rightarrow \ci{\set{X}}{\set{Y}}{\set{Z}\cup \set{W}}$ & \\
  & & \\
	\textbf{Intersection} & $\ci{\set{X}}{\set{Y}}{\set{Z}~\cup~\set{W}}~\&~\ci{\set{X}}{\set{W}}{\set{Z}~\cup~\set{Y}} \Rightarrow \ci{\set{X}}{\set{Y}~\cup~\set{W}}{\set{Z}}$, & \\

  & & \\

  \end{tabular}

where $\set{X}$, $\set{Y}$, $\set{Z}$ and $\set{W}$ are all disjoint subsets 
of the set of all the variables in the domain $\set{V}$, and $\lambda$ stands for a single variable, not in 
$\set{X} \cup \set{Y} \cup \set{Z} \cup \set{W}$.
The intersection axiom is valid only for strictly positive probability distributions.
This list of axioms represents the relationships that hold among the independences encoded by the graph. 

In summary, when the distribution $P$ is a graph-isomorph, there exists a graph $G$ that is a perfect-map for $P$.
For representing a distribution $P$, any graph $G$ which is an I-map of $P$ may be used.
However, the more independences of the underlying distribution encoded in the graph, 
the better the model is in complexity and accuracy when used for inference.
Assuming graph-isomorphism is an important decision,  
since not all the existent distributions may be represented by an undirected graph.
For example, there are distributions that may be represented by an acyclic directed graph,
and in this case, Bayesian networks are the correct model to use. There are also other distributions
that cannot be encoded by a graph.

\subsubsection{The Markov blanket concept \label{sec:blanket}}
This section describes the concept of Markov blanket, 
a central theoretical concept in the representation of distributions,
introduced by \cite{pearl88}.
The Markov blanket of a variable is the only knowledge needed to predict 
the behavior of that variable. Hence, this concept holds relevance 
for a wide variety of applications where
local relationships to some variables are significant.

\begin{definition}{The Markov blanket concept}.

The Markov blanket of a variable $X$
is a minimal set, denoted here $\blanket{X}$, conditioned on which all other nodes in the domain of variables $\set{V}$ are 
independent of $X$, that is, 
\begin{equation}
 \forall Y \in \set{V}-\{\blanket{X}\},~\ci{X}{Y}{\blanket{X}}.
\end{equation}
\end{definition}

That is, the Markov blanket of a variable 
is the smallest set of variables that shields it from the probabilistic influence 
of the variables not in the blanket. 
From a graphical view point, the Markov blanket of a variable $X$
is identical to  its neighbors in the graph. 

In the textbook of \cite{pearl88}, it is proved formally 
that, for strictly positive distributions, the independence structure can be constructed by piecing together 
the Markov blanket of all the variables of the domain,
connecting with an edge every two variables $X$ and $Y$, such that $X$ belongs to the Markov blanket of $Y$. 
There is also a proof stating that every variable $X \in \set{V}$ in a distribution that is a graph isomorph, and therefore satisfies the
Pearl's axioms, has a unique Markov blanket. 
As only strictly positive distributions satisfy the Intersection axiom, this mechanism for constructing the structure 
only holds for positive distributions.

\subsection{Parameterization}
This section explains how to quantify the relationships encoded in $G$.
Although this work only addresses the problem of structure learning, 
the quantitative aspects of Markov networks are briefly explained 
for better clarifying our work.
Below is described a factorization method for constructing 
the Gibbs distribution for an arbitrary undirected graph $G$,
provided in \cite{pearl88}:

\begin{itemize}
 \item Identification of the maximal subgraphs whose nodes are all adjacent to each other, 
called the \emph{maximal cliques} of $G$. 
For example, the graph in Figure~\ref{fig:undirectedgraphs}~(a) shows a maximal clique of size $4$ 
among the nodes corresponding to variables $\{7,8,10,11\}$, 
two maximal cliques of size $3$ among nodes $\{2,3,5\}$ and $\{8,9,11\}$, 
and the rest of edges are maximal cliques of size $2$. 
In Figure~\ref{fig:undirectedgraphs}~(b) the size of all the cliques is $2$.

 \item For each clique $\set{c} \in \mathcal{C}$ in the set of all the cliques in the graph, 
assign a non-negative potential function $g_{\set{c}}(\set{X}_\set{c})$ 
(where $\set{X}_\set{c}$ is the set of variables that belong to the clique $\set{c}$)
measuring the relative degree of compatibility associated with each possible configuration of $\set{X}_\set{c}$.
Usually each potential function is represented 
by a table with a numerical parameter assigned 
to each possible complete assignment of the variables that compose the clique,
like the tabular model showed in Figure~\ref{fig:tabularmodel},
but including only the variables that compose the clique $\set{c}$.
A difference with the tabular model is that here the parameter values are not normalized. 

 \item Form the product $\prod\limits_{\set{c} \in \mathcal{C}}{g_{\set{c}}(\set{X}_\set{c})}$ 
of the potential functions over all the cliques.

 \item Construct the Gibbs distribution by normalizing the product over all possible value combinations 
of the variables in the system

\begin{equation}  \label{eq:jointdist}
 P(X_0,..,X_{n-1}) = \frac{1}{Z} \prod\limits_{\set{c} \in \mathcal{C}}{g_{\set{c}}(\set{X}_\set{c})},
\end{equation}
where Z is the \emph{partition function}, or \emph{normalization constant}, computed as
\begin{equation}\label{eq:partitionfunction}
 Z = \sum\limits_{X_0,..,X_{n-1}}{\prod\limits_{\set{c} \in \mathcal{C}}{g_{\set{c}}(\set{X}_\set{c})}}.
\end{equation}
\end{itemize}

Using the Hammersley-Clifford theorem it is possible to prove that
the general form of the Gibbs distribution of Equation~(\ref{eq:jointdist}) 
embodies all the conditional independences  
encoded in the graph $G$. 
Such form of the Gibbs distribution presents some difficulties.
First, it is difficult to discern the meaning of the potential functions. 
Second, the computational cost of calculating the partition function $Z$ is exponential, 
as it requires an exponential sum over all possible assignments of the complete set of variables.

% \newpage
\section{The Markov networks learning problem\label{sec:structurelearning}}

This section discusses the difficulties that arise in the task 
of learning Markov networks from historical information.
This task is only possible whenever the size of the input dataset $D$ is sufficient, 
and the data is a representative sample of the underlying distribution $P$.
When these conditions are satisfied, it is possible that some algorithms 
learn a model for representing $P$ by exploring and analyzing $D$.
The input dataset $D$ contains historical information commonly structured in the \emph{tabular format}, 
a standard format in machine learning.
This is a file that contains a table with a column per random variable in $P$, 
and the rows are the datapoints, each one being a complete assignment for all the variables. 
For example, a datapoint for a domain with $n=4$ random binary variables
$\set{V}=\{X_0,X_1,X_2,X_3\}$ 
may be $(X_0=0,X_1=1,X_2=1,X_3=0)$. 
The algorithms discussed in this work ignore the problem of missing values,
which is solved by known, yet computationally challenging, statistical techniques.
% Missing data is not allowed, and in general a pre-processing of the input dataset may be necessary to deal with this. 

Learning a Markov network from data is a problem that consists in learning both 
the structure $G$ and the parameters $\parameters$.
Of course, the best possible structure learned is a perfect-map, 
that is, a model that contains a structure encoding all the dependences and the independences present in $P$.
However, every model containing a structure which is an I-map of $P$ is a good solution.
The closer to a perfect-map, the better is the structure learned, 
and the better is the resulting Markov network for representing $P$.
When learning a model for large domains, a desirable property of the model is the sparsity,
since densely connected models require too many parameters,
and make exact and even approximate inferences computationally intractable.

\subsection{Goals of Markov networks learning \label{sec:goalsoflearning}}
For evaluating the merits of a model learning method, it is important to
consider the goal of learning. 
Clearly, learning the complete model (structure plus parameters)
is the ideal method, but due to computational, spatial or sampling limitations,
it may not be possible in practice. For that reason, other less ambitious goals 
are often considered in practice, such as the three main goals of learning discussed by \cite{koller09}:

	\begin{itemize}
	\item \textbf{Density estimation: }
	A common reason for learning a Markov network is to use it for some inference task.
	When formulating the goal of learning as one of density estimation, the goal is
	to construct a model $M$ so that the defined distribution 
	is ``close'' to the underlying distribution $P$. 
	A common metric for evaluating the quality of such approximation is 
	the use of the likelihood of the data $Pr(D\mid M)$. 
	However, this goal assumes that the overall distribution $P$ is needed. 

	\item \textbf{Specific prediction tasks: }
	The goal is predicting the distribution of a particular set of variables
	$\set{Y}$, 
	given certain set of variables $\set{X}$. 
	When the model is used only to perform a particular task, 
	if the model is never evaluated on predictions of the variables $\set{X}$, 
	it is better to optimize the learning task for improving the quality of its
	answers to $\set{Y}$. 
	This has been the goal of a large fraction of the work in machine learning. 
	For example, consider the problem of documents classification for a given set
	of relevant words of a document, 
	and a variable that labels the topic of the document. 
	Another well-known example is the task of image segmentation, where the goal of the task 
	is the prediction of class labels for all the pixels in the image, given the
	image features. 

	\item \textbf{Knowledge discovery:} 
	The goal of knowledge discovery is to learn the correct structure of the underlying distribution. 
	There are some cases when the learned structure can reveal some important unknown properties
	of the domain. 
	It is a very different motivation for learning the distribution. 
	An examination of the learned structure can show dependences among variables 
	as positive or negative correlations.
	In a knowledge discovery application, it is far more critical to assess the
	confidence in a prediction,
	taking into account the extent to which it can be identified given the available
	data, and the number of hypotheses 
	that would cause similar observed behavior. 
	For example, in a medical diagnosis domain, we may want to learn the structure
	of the model 
	to discover which predisposing factors lead to certain diseases, and which
	symptoms are associated with different diseases. 

	\end{itemize}

\subsection{Parameters estimation \label{sec:parameterization}}

Markov network parameters estimation is usually used 
to choose the value of the parameters by fitting the model to data,
because tuning parameters manually is often difficult, 
and learned models often exhibit better performance.
This task has shown to be an NP-hard problem by \cite{BARAHONA82}.
 
For estimating the parameters, the most common method proposed is \emph{maximum-likelihood estimation}, 
potentially using some regularization as an additional parameter prior.
Unfortunately, evaluating the likelihood of a complete model requires, 
for every set of parameters proposed during the maximum-likelihood estimation process, 
the computation of the partition function
$Z$, which is used for normalizing the product over all possible value combinations of the variables of the domain,
as showed in Equation~(\ref{eq:partitionfunction}). 
Although it is not possible to optimize the maximum-likelihood in a closed form,
it is guaranteed that the global optimum can be found, 
because it is a concave function.
As a result, some approximations and heuristics in the literature are introduced in \cite{Minka_2001,Vishwanathan2006},
for reducing the cost of parameters estimation,
using iterative methods such as simple gradient ascent,
or other sophisticated optimization algorithms.
Unfortunately, this problem remains intractable in practice, 
because the use of the partition function 
couples all the parameters across the network,
requiring several inference steps on the network 
(iterative methods with interleaved inference). 
 
For reducing the cost of parameters estimation, other solutions have been proposed. 
Pseudolikelihood by \cite{besag77}, and Score Matching by \cite{Hyvarinen05}
are some tractable approximate alternatives. 
The \emph{loopy belief propagation} method proposed in \cite{pearl88} and later in \cite{yedidia2005},
and some variants introduced in \cite{Wainwright2008}, 
uses an approximate inference technique for approximating 
the gradient of the maximum likelihood function. 
Another solution for outperforming the robustness of loopy belief propagation is provided in \cite{ganapathi2008}.

For avoiding overfitting, many of these scoring methods 
commonly need the use of a regularization term adding an extra hyper-parameter, 
whose best value has to be found empirically.
For example, it can be found by running the training stage for several values
of the hyper-parameter, potentially using cross-validation.

\subsection{Structure learning approaches}
The two broad approaches for learning the structure of Markov networks from data are  
\emph{score-based} and \emph{independence-based} approaches. 
The former is intractable in practice, and the latter is efficient but presents quality problems.
Both approaches have been motivated by distinct learning goals (those described in Section~\ref{sec:goalsoflearning}).
Generally, score-based approaches may be better suited for the density estimation goal, that is, 
tasks where inferences or predictions are required.
As explained below in Section~\ref{sec:scorebased}, score-based methods 
learn the complete Markov network (structure and parameters).
There is an overwhelmingly use of Markov networks for such settings, 
including image segmentation and others, where there exists a particular inference task in mind.
Instead, independence-based methods are better suited for the remaining goals, that is, 
for specific prediction tasks, and knowledge discovery.
On one hand, independence-based algorithms are commonly used for tasks such as feature selection for classification,
since it is possible to perform local discovery for a particular set of variables of interest 
(more details in Section~\ref{sec:independenBased}).
On the other hand, independence-based algorithms are suited for knowledge discovery tasks,
that is, tasks where understanding the interactions among variables in
a domain carries the greatest importance, or where the structure is viewed purely as a predictive tool,
for example, econometrics, psychology, or sociology. 

Since this work focuses on the independence-based approach 
to Markov networks structure learning methods, Sections~\ref{sec:approaches} and~\ref{sec:analysis}
only discuss in detail the state-of-the-art independence-based algorithms.

\subsubsection{Score-based approach \label{sec:scorebased}}
Score-based algorithms were proposed for learning the structure of Bayesian networks,
in the works of \cite{LamBacchus94} and \cite{Heckerman95}, 
and later proposed for learning the structure of Markov networks, 
in the works of \cite{PietraPL97} and \cite{MCCALLUM03}.
Such algorithms approach the problem as an optimization over the space of complete models, 
looking for the one with maximum \emph{score}. 
The goal of score-based algorithms is to find the model that maximizes its score.
Traditional score-based algorithms perform a global search to learn a set of potential functions
that accurately captures high-probability regions of the instance space of complete models. 

The standard approach for learning the structure of Markov networks
with a score-based approach is the Della Pietra et al. algorithm.
This algorithm learns the structure by inducing a set of potential functions from data.
Its strategy is based on a top-down search, that is, a general-to-specific search.
This algorithm starts with a set of atomic potentials (that is, exclusively the variables of the domain).
Then, it creates a set of candidate potentials in two ways. 
First, each potential currently in the model 
is conjoined (i.e., associated) with every other potential in the model.
Second, each potential in the model is composed with each atomic potential.
Then, for efficiency reasons, the parameters are learned for each candidate potential, assuming
that the parameters of all other potentials remain unchanged.
When setting the parameters, it uses the Gibbs sampling for inference.
Then, for each candidate potential, the algorithm 
evaluates how much adding such potential would increase the log-likelihood, 
which is the score used by this algorithm.
The potential that maximizes this measure is added.
When no one candidate potential improves the score of the model, the procedure ends.
Another algorithm using the same approach 
is proposed in \cite{MCCALLUM03}.
It is a similar algorithm to the one proposed by Della Pietra, but
performing an efficient heuristic search over the space of candidate structures,
for automatically inducing potentials that most improve the conditional log-likelihood.
However, as reported by \cite{DavisAndDomingos2010:BottomUp}, such general-to-specific 
search methods are inefficient, because 
they test many potential variations with no support in the data, 
and because they are highly prone to local optima.

Recently, other alternative approaches have been considered.
The approach of \cite{Lee+al:NIPS06}, ~\cite{Hofling09}, and
~\cite{ravikumar2010:l1} 
propose to couple parameters learning and potentials induction into one step, 
by using $L_1$-regularization, which forces most numerical parameters to be zero.
They approach the problem as an optimization problem,
providing a large initial potential set,
with all the possible potentials of interest.
Then, after learning, model selection occurs by selecting 
those potentials with non-zero parameters.
For efficiency reasons, the approaches of H{\"o}fling and Tibshirani,
and Ravikumar et al., only construct pairwise networks 
(networks involving only cliques of size two or one for factorization).
Instead, the algorithm of Lee et al., can learn arbitrarily long potentials.
In practice, however, it has been evaluated only for inducing potentials of length two
(that is, for learning pairwise networks).

A recent alternative approach was proposed by \cite{DavisAndDomingos2010:BottomUp},
called the Bottom-up Learning of Markov Networks (BLM) algorithm.
BLM starts with each complete training example as a long potential in the Markov network.
Then, the algorithm iterates through the potential set,
generalizing each potential to match its k-nearest previously unmatched examples by dropping variables.
When the new  generalized potential improves the score of the model, it is incorporated to the model.
The loop ends when no generalization can improve the score.

However, all these approaches are often slow for two reasons.
First, the size of the search space of structures is intractable in the number of variables.
Second, for evaluating the score at each step, it is necessary to compute the score,
requiring the estimation of the numerical parameters (an NP-hard task, 
as explained in Section~\ref{sec:parameterization}).

\subsubsection{Independence-based approach \label{sec:independenBased}}

Independence-based (also known as constraint-based) algorithms work by performing a succession 
of \textit{statistical independence tests} 
for discovering the independence structure of graphical models. 
These algorithms exploit the semantics of the independence structure,
casting the problem of structure learning as an instance of the constraint satisfaction problem,
where the constraints are the independences present in the input dataset 
(and therefore, in the underlying distribution), 
and the goal is to find a structure encoding all such independences.

Each independence test consults the data for responding to a query about the
conditional independence among some input random variables $X$ and $Y$, 
given some conditioning set of variables $\set{Z}$, 
resulting in an independence assertion $\ci{X}{Y}{\set{Z}}$, 
or $\cd{X}{Y}{\set{Z}}$ for a dependence assertion.
The computation cost of statistical tests is proportional
to the number of rows in the input dataset $D$, and the number of variables involved.
Examples of independence tests used in practice are Mutual Information
in \cite{cover&tomas91}, Pearson's $\chi^2$ and $G^2$ in \cite{AGRESTI02}, 
the Bayesian test in \cite{MARGARITIS05}, and for continuous Gaussian data 
the \emph{partial correlation} test in \cite{Spirtes00}.
Such independence tests compute a statistical value 
for a triplet of variables $\triplet{X}{Y}{\set{Z}}$, given an input dataset, and decide
independence or dependence comparing it with a threshold.
For instance, $\chi^2$ and $G^2$ use the \emph{p-value}, 
which is computed as the probability of obtaining a test statistic 
at least as extreme as the one that was actually observed, 
assuming that the null hypothesis is true (that is, variables are dependent). 
The null hypothesis is rejected when the p-value is less than the significance 
level $\alpha$, which is often $0.05$ or $0.01$. 
When the null hypothesis is rejected, the result is said to be statistically significant.

An elegant, efficient and scalable strategy used by several independence-based algorithms 
in the literature is called the \emph{local-to-global} strategy, presented
in a recent work of \cite{Aliferis2010b}.
This is a generalization of previous algorithms using such strategy.
Algorithm~\ref{algLGLMN} shows the outline of this theoretically sound and straightforward procedure,
omitting the third step of edges orientation, used for learning Bayesian networks . 

\begin{algorithm}[ht!]                      
\caption{LGL for Markov networks}           
\label{algLGLMN}                           
\begin{algorithmic}[1]                    
\STATE \label{stepFindMBs} Learn $\blanket{X_i}$ for every variable $X_i\in \set{V}$.
\STATE Piece-together the global structure using an ``OR rule''.
\end{algorithmic}
\end{algorithm}

Such a strategy suggests to construct the independence structure
by dividing the problem into $n$ different Markov blanket learning problems, that is, 
the Markov blanket is learned for each variable of the domain $\set{V}$.
The learning of Markov blanket is generalized by Aliferis et al., for learning Bayesian networks,
in the Generalized Local Learning (GLL) framework \cite{Aliferis2010}. 
Algorithms using a local-to-global strategy learn locally the Markov blanket of every variable in the domain, 
and then construct a global structure linking each of these variables with every member of its Markov blanket, 
using an ``OR rule'' (an edge exists between two variables $X$ and $Y$ when $X\in\blanket{Y}~or~Y\in\blanket{X}$).

For learning the Bayesian networks structure, independence-based algorithms first arose in 1993,
when \cite{Spirtes00} published the well-known algorithms SGS and PC,
in the first edition of such textbook. 
Then, other independence-based algorithms appeared in works about feature selection 
via the induction of Markov blanket, and works about Bayesian and Markov networks structure learning.
For that reason, a series of independence-based algorithms for Markov blanket learning of Bayesian networks appeared, 
such as the Koller-Sahami (KS) algorithm in \cite{KollerSahami1996:KS}, the Grow-Shrink (GS) algorithm in \cite{MARGARITIS00}, 
the Incremental Association Markov Blanket (IAMB) algorithm and its variants in \cite{tsamardinos03:IAMB} , 
the Max-Min Parents and Children Markov Blanket (MMPC/MB) algorithm in \cite{Tsamardinos2006:MMHC}, 
the HITON-PC/MB algorithm in \cite{Aliferis2003:HITON}, 
the Fast-IAMB algorithm in \cite{Yaramakala2005:fastIAMB},
the Parent-Children Markov Blanket (PCMB) algorithm in \cite{penia2007:PCMB}
and the Iterative Parent and Children Markov Blanket (IPC-MB) in \cite{Fu2008:IPCMB}.
A summary of the most important aspects of such algorithms 
is shown in Table~\ref{tab:blanketalgs}, reproduced from  
the conclusions of a recent review of Markov blanket based feature selection 
written by \cite{fu2010:reviewMB}.

\begin{table}[ht]
\scriptsize
\center
	      \caption{Summary of Markov blanket learning algorithms for Bayesian networks. \label{tab:blanketalgs}}
		      \begin{tabular}{p{1.5cm}  p{3cm}  p{8cm}}
			    \multicolumn{3}{c}{~}\\
			    \hline
			    \\
			    \textbf{Name} & \textbf{Reference} & \textbf{Comments} \\ 
			    \\ \hline
			    \\
			    KS & \cite{KollerSahami1996:KS}  & \begin{minipage}{10cm}
					\begin{itemize}% \setlength{\itemsep}{-2mm} 
					\item Not Sound
					\item The first one of this type
					\item Requires specifying MB size in advance
					\end{itemize}
					\end{minipage} \\ 
			    \\ \hline
			    \\   
			    GS & \cite{MARGARITIS00} & \begin{minipage}{10cm}
					\begin{itemize} % \setlength{\itemsep}{-2mm} 
					\item Sound in theory
					\item Proposed to learn Bayesian network via the induction \\ of neighbors of each variable
					\item First proved such kind of algorithm
					\item Works in two phases: grow and shrink
					\end{itemize}
					\end{minipage} \\ 
			    \\ \hline
			    \\
			    IAMB and its variants & \cite{tsamardinos03:IAMB} & \begin{minipage}{10cm}
					\begin{itemize} % \setlength{\itemsep}{-2mm} 
					\item Sound in theory
					\item Actually variant of GS
					\item Simple to implement
					\item Time efficient
					\item Very poor on data efficiency
					\item IAMB's variants achieve better performance on data \\ efficiency than IAMB
					\end{itemize}
					\end{minipage} \\ 
			    \\ \hline
			    \\
			    HITON-PC/MB & \cite{Aliferis2003:HITON} & \begin{minipage}{10cm}
					\begin{itemize} % \setlength{\itemsep}{-2mm} 
					\item Not sound
					\item Another trial to make use of the topology \\ information to enhance data efficiency
					\item Data efficiency comparable to IAMB
					\item Much slower compared to IAMB
					\end{itemize}
					\end{minipage} \\ 
			    \\ \hline
			    \\
			    Fast-IAMB & \cite{Yaramakala2005:fastIAMB} & \begin{minipage}{10cm}
					\begin{itemize} % \setlength{\itemsep}{-2mm} 
					\item Sound in theory
					\item No fundamental difference as compared to IAMB
					\item Adds candidates more greedily to speed up the learning
					\item Still poor on data efficiency performance
					\end{itemize}
					\end{minipage} \\ 
			    \\ \hline
			    \\
			    MMPC/MB & \cite{Tsamardinos2006:MMHC} & \begin{minipage}{10cm}
					\begin{itemize} % \setlength{\itemsep}{-2mm} 
					\item Not sound
					\item The first to make use of the underling topology information
					\item Much more data efficient compared to IAMB
					\item Much slower compared to IAMB
					\end{itemize}
					\end{minipage} \\ 
			    \\ \hline
			    \\
			    PCMB & \cite{penia2007:PCMB} & \begin{minipage}{10cm}
					\begin{itemize} % \setlength{\itemsep}{-2mm} 
					\item Sound in theory
					\item Data efficient by making use of topology information
					\item Poor on time efficiency
					\item Distinguish spouses from parents/children
					\item Distinguish some children from parents/children
					\end{itemize}
					\end{minipage} \\ 
			    \\ \hline
			    \\
			    IPC-MB & \cite{Fu2008:IPCMB} & \begin{minipage}{10cm}
					\begin{itemize} % \setlength{\itemsep}{-2mm} 
					\item Sound in theory
					\item Most data efficient compared with previous algorithms
					\item Much faster than PCMB on computing
					\item Distinguish spouses from parents/children
					\item Distinguish some children from parents/children
					\item Best trade-off among this family of algorithms
					\end{itemize}
					\end{minipage} 
			    \\ 
			    \\ \hline

			\end{tabular}

\end{table}

For learning the Markov networks structure, independence-based algorithms arose later in 2006,
when \cite{Bromberg06,brombergmargaritis09b} published the Grow-Shrink Markov Network (GSMN) algorithm 
and the Grow-Shrink Inference-based Markov Network (GSIMN) algorithm.
Then, other independence-based algorithms appeared for Markov networks structure learning, 
such as the Particle Filter Markov Network (PFMN) algorithm in \cite{BROMBERG07,margaritisBromberg09}, 
and the Dynamic Grow Shrink Inference-based Markov Network (DGSIMN) algorithm in \cite{gandhi08}.
Another approach is proposed in \cite{Bromberg08:thesis,BrombMarg09}, as a framework based on argumentation 
for improving reliability of tests.
In Section~\ref{sec:approaches}, all these independence-based algorithms 
for learning the structure of Markov networks are surveyed in detail.

There are several advantages of independence-based algorithms. 
First,  they can learn the structure 
without interleaving the expensive task of parameters estimation 
(contrary to score-based algorithms, as explained before), 
reaching sometimes polynomial complexities in the number of statistical tests performed. 
If the complete model is required, the
parameters can be estimated only once for the given structure. 
Another important advantage of such algorithms is that they are sound, 
that is, when statistical tests outcomes are correct, 
the structure found correctly represents the underlying distribution.
However, they are correct under the following assumptions: 
\begin{itemize}
 \item the distribution of data is a \emph{graph-isomorph}
 \item the underlying distribution is strictly positive 
 \item the outcomes of tests are reliable
\end{itemize}

The third condition for soundness is an important problem of independence-based algorithms. 
When the dataset used for learning is not sufficiently large, 
or it is not a representative sampling of the underlying distribution,
the outcomes of tests are incorrect, and the structures learned are deemed unreliable.
This problem of statistical tests unreliability is exponentially exacerbated  with the
number of variables involved (for some fixed size of dataset).
For good quality, statistical tests require enough counts in their contingency
tables, and there are an exponentially number of those 
(one per value assignment of all variables in the test). 
For example, \cite{COCHRAN54} recommends that the $\chi^2$ test 
must be deemed unreliable when more than $20\%$ of these cells 
have an expected count of less than 5 data points.

Another disadvantage of independence-based algorithms is that 
there is no guarantee about the quality of the complete model obtained by learning 
the structure first, and then fitting parameters for such learned structure.
This is an approximation, and there are no experimental results published in the literature 
about independence-based methods for learning complete models.

% \newpage
\section{Independence-based algorithms for learning the Markov networks structure \label{sec:approaches}}
This section reviews the independence-based structure learning algorithms 
for Markov networks that have appeared in the literature. 
The review on this section covers a series of published algorithms that tackle such problem. 

\subsection{The Grow-Shrink Markov Network algorithm \label{sec:gsmn}}
The Grow-Shrink Markov Network (GSMN) algorithm was introduced 
in \cite{Bromberg06,brombergmargaritis09b}, as the first independence-based structure learning algorithm 
for Markov networks in the literature.
This algorithm is an adaptation to Markov networks of the GS algorithm of \cite{MARGARITIS00}
for learning the Markov blanket.

The GSMN algorithm learns the global structure of a Markov network
following the simple outline of local-to-global algorithms showed in Algorithm~\ref{algLGLMN},
and using the GS algorithm outlined in Algorithm~\ref{alg:gs} for discovering the Markov blanket of the variables.
\begin{algorithm}[h!t]  
  \caption{\label{alg:gs} $\mathbf{GS}(X, \set{V})$.}
  \begin{algorithmic}[1]
    \STATE $\set{S} \longleftarrow \emptyset. $ \label{line:line1}
    \STATE sort $\set{V}-\{X\}$ by increasing association with $X$ \label{line:line2}
    \medskip
    \STATE /* Grow phase */
    \STATE $\mathbf{while}~\exists Y \in \set{V}-\{X\} \mbox{~s.t.~}\cd{Y}{X}{~\set{S}},~\mathbf{do}~ \set{S} \leftarrow \set{S}  \cup \{Y\}$. \label{line:growGS}
    \medskip
    \STATE /* Shrink phase */
    \STATE $\mathbf{while}~\exists Y \in \set{S} \mbox{~s.t.~}\ci{Y}{X}{~\set{S}-\{Y\}},~\mathbf{do}~ \set{S} \leftarrow \set{S} - \{Y\}.$ \label{line:shrinkGS}
%     \medskip
    \RETURN $\set{S}$ \label{line:endGS}
  \end{algorithmic}
\end{algorithm}
GS maintains a set called $\set{S}$ (initialized empty in line~\ref{line:line1})
that contains the Markov blanket of the input variable $X$ when the algorithm terminates.  
First, in line~\ref{line:line2}, GS performs an initialization phase that sorts by increasing association 
with $X$ the rest of the variables of the domain $\set{V}$ ,
using an unconditional test between $X$ and every variable $Y \in \set{V}-\{X\}$.
Then, the algorithm proceeds in two stages, the \emph{grow} and \emph{shrink} phases, using such ordering. During
the grow phase (line \ref{line:growGS}) the algorithm increases the set
$\set{S}$ with every variable $Y$ that is found dependent on $X$
conditioning on the current state of $\set{S}$. 
By the end of this phase, the set $\set{S}$  
contains all members of the Markov blanket, 
but potentially includes some false positives that are non-members.
These false positives are removed during the shrink
phase (line \ref{line:shrinkGS}), where variables found independent of $X$
conditioning on the set $\set{S}$  are removed from $\set{S}$.

The main advantages of GSMN are 
\textit{i)} it is sound, and \textit{ii)} it is efficient.
The soundness of GSMN is proven theoretically by its authors,
guaranteeing that a correct independence structure is found when statistical tests are reliable.
This algorithm is efficient because it is polynomial in the number of independence tests for discovering the structure,
each test requiring a polynomial time execution with respect to the variables involved in the test, and the size of the input dataset.
A disadvantage of using GS is that unreliable statistical tests produce cascade errors,
not only with incorrect outcomes, but that also generate next incorrect tests during grow and shrink phases, 
producing cumulatively errors, as stated in \cite{Spirtes00}. 

Two other important algorithms for learning the Markov blanket of a variable, for Bayesian networks,
are the Incremental Association Markov Blanket (IAMB) algorithm in \cite{tsamardinos03:IAMB}, and 
the HITON algorithm in \cite{Aliferis2003:HITON}.
Both algorithms have been proven empirically to be more resilient
than GS to the errors of statistical tests, by introducing two simple variants. 
On the one hand, the IAMB algorithm only introduce a modification
by interleaving the initialization step of ordering in the grow phase 
(i.e., interleaves lines~\ref{line:line2} and \ref{line:growGS} of Algorithm~\ref{alg:gs}).
By interleaving the sorting step in the grow phase, 
IAMB maximizes the accuracy, reducing the number of false positives in the grow phase.
On the other hand, the HITON algorithm aims to reduce the data requirements of IAMB,
by introducing an additional modification in the criteria used for testing independence.
In both grow and shrink phases, instead of only conditioning on its tentative Markov blanket $\set{S}$, 
HITON tests independence conditioning in any of the subsets of $\set{S}$ (that is, every set $\set{Z} \subseteq \set{S}-\{Y\}$). 
As statistical tests are more reliable while containing fewer variables, 
such modification exploits the Strong union axiom of Pearl, for 
improving the quality of independence tests when data is scarce.
A disadvantage of the approach proposed by HITON 
is its exponential cost in $\mid\set{S}\mid $ (i.e., the size of $\set{S}$) ,
but in general $\mid\set{S}\mid$ is comparatively smaller than the size of the domain~$n$. 
In summary, both algorithms are proven to be better in quality than GS, 
but they were designed for learning the structure of Bayesian networks,
and there are not any works in the literature proposing 
a theoretical adaptation of such ideas for learning the complete structure of a Markov network,
or empirically evaluating its performance. 

\subsection{The Grow Shrink Inference Markov Network algorithm} 

The Grow Shrink Inference Markov Network (GSIMN) algorithm was presented in \cite{Bromberg06,brombergmargaritis09b}. 
This algorithm works in a similar fashion to that of GSMN algorithm, using the  
local-to-global strategy of Algorithm~\ref{algLGLMN},
and learning the Markov blanket of all the variables with the GS algorithm,
but interleaving an inference step to reduce the number of tests required to learn the Markov blanket.
By using a theorem for inference called the Triangle theorem by the authors, 
GSIMN reduces the number of tests performed on data, 
without adversely affecting the quality of the learned structures.
It may be useful when using large datasets, or in distributed domains, 
where statistical tests are very expensive.

GSIMN introduces the Triangle theorem, 
based on the Pearl's axioms showed in Section~\ref{sec:correctness}.
This is a sound theorem for allowing to infer unknown independences from those known so far.

\begin{theorem}[Triangle theorem]
  Given Eqs.~(\ref{eq:generalaxioms}), for every variable $X$, $Y$, $W$ and
  sets $\set{Z}_1$ and $\set{Z}_2$ such that
  $\{X,Y,W\}\cap\set{Z}_1=\{X,Y,W\}\cap\set{Z}_2=\varnothing$,
\begin{eqnarray*}
  \cd{X}{W}{\set{Z}_1} \wedge \cd{W}{Y}{\set{Z}_2}
    & \implies &
    \cd{X}{Y}{\set{Z}_1 \cap \set{Z}_2} \\
  \ci{X}{W}{\set{Z}_1} \wedge \cd{W}{Y}{\set{Z}_1 \cup \set{Z}_2}
    & \implies &
    \ci{X}{Y}{\set{Z}_1}.
\end{eqnarray*}
The first relation is called the ``D-triangle rule'' and the second
the ``I-triangle rule.''
\end{theorem}

When GSIMN tests some independence on data, 
first applies the Triangle theorem to the tests already
done on data, in order to check if such independence assertion can be logically inferred. 
If the test cannot be inferred, then this is done on data, and stored. 
For convenience, the algorithm determines the visit ordering (the order for local learning) 
in an attempt to maximize the use of inferences.
The results obtained with GSIMN show savings up to a $40\%$ in the running times of GSMN,
obtaining comparable qualities.

\subsection{Particle Filter Markov networks algorithm \label{sec:pfmn}}
The Particle Filter Markov networks algorithm (PFMN) is presented in \cite{BROMBERG07,margaritisBromberg09}, 
as a novel independence-based approach for learning Markov network structures.
Previous independence-based algorithms reviewed, such as the GSMN and GSIMN, 
use the local-to-global strategy. Instead, 
this algorithm learns directly a global structure as the solution.

PFMN was designed for improving the efficiency of the GSIMN algorithm.
This algorithm works performing statistical independence tests iteratively,
by selecting greedily at each iteration the statistical test that eliminates 
the major number of inconsistent structures.
This decision is taken by first modeling the learning problem with a Bayesian approach,
selecting as the solution the structure $G$ that maximizes its posterior probability $\Pr(G\mid D)$.
Since the direct computation of such probability is intractable, 
PFMN propose a generative model with independence tests which is an approximation to that posterior probability.
With this model, it is possible to compute efficiently such probability, given the information over a set of independences.
Moreover, the authors claim that it is possible to demonstrate that, under the assumption of correctness of tests, 
the distribution of $\Pr(G\mid D)$ converges to a correct structure. 
% Because of this, the algorithm utilizes a heuristic for choosing greedily the test which 
% minimizes the expected entropy (or maximizes the information gain).

This approach is useful in domains where independence tests are expensive, such as cases
of very large data sets or in distributed domains. 
Results obtained by PFMN show improvements in running times up to $90\%$ with respect to GSIMN,
and comparable qualities on structures found by GSIMN and GSMN.

\subsection{The Dynamic Grow Shrink Inference-based Markov Network algorithm}

The Dynamic Grow Shrink Inference-based Markov Network (DGSIMN) algorithm was presented in \cite{gandhi08}. 
This is an extension of the GSIMN algorithm which, in the same way than GSIMN, 
uses the Triangle theorem for avoiding unnecessary tests. 
The outline of DGSIMN is similar to GSMN and GSIMN, using the  
local-to-global strategy of Algorithm~\ref{algLGLMN},
and the GS algorithm showed in Algorithm~\ref{alg:gs} for learning the Markov blanket of the variables,
but interleaving a different inference step than GSIMN
for reducing the number of tests performed.

DGSIMN improves the GSIMN algorithm by dynamically selecting the locally optimal 
test that will increase the state of knowledge about the structure, 
estimating the number of inferred independences that will be obtained after executing a test, 
and selecting the one that maximizes such number of inferences. 
This helps decreasing the number of tests required to be evaluated on data, 
resulting in an overall decrease in the computational requirements of the algorithm.

The results of experiments with the DGSIMN algorithm shows that it improves the fixed ordering 
of variables in the Markov blanket learning subroutine,
improving the running times of GSIMN up to $85\%$, 
obtaining comparable qualities to GSMN.

\subsection{Argumentation for improving reliability \label{sec:argumentation}}
Algorithms presented in previous sections are independence-based algorithms 
that focus on improving the efficiency, ignoring the important problems in the quality of learned structures
that arises when statistical tests are not reliable, due to data scarceness.

An independence-based approach for dealing with unreliable tests 
was presented in \cite{Bromberg08:thesis,BrombMarg09}, 
by modeling the problem of low reliability of independence tests as a knowledge base 
with independence assertions that may contain errors due to incorrect statistical tests performed,
and the Pearl's axioms (directed or undirected axioms, depending on the target model to learn).
The advantage of this approach is its power for correcting errors of tests 
by exploiting logically the independence axioms of Pearl.
When exist independence assertions in the knowledge base that are in conflict,
it is clear that some independence assertions are incorrect,
and this approach proposes to resolve such conflicts through the \emph{argumentation} framework, 
which is a defeasible logic proposed by \cite{Amgoud2002:arg}, to reason about and correct errors.

This approach was presented as a more robust conditional independence test called 
the \emph{argumentative independence test}, for learning both Bayesian and Markov networks. 
Experimental evaluation shows significant improvements in the accuracy 
of the argumentative independence test over other simple statistical tests (up to $13\%$),
and improvements on the accuracy of Blanket discovery algorithms such as PC and GS (up to $20\%$).

A disadvantage with this approach is that, as it is a propositional formalism,
it requires to propositionalizing the set of rules of Pearl, which are first-order.
As these are rules for super-sets and sub-sets of variables, 
its propositionalization involves an exponential number of propositions,
and then, the exact argumentative algorithm proposed is exponential.
In this work, an approximate solution is presented with polynomial
running time, still improving the quality in the experimental evaluation (up to $9\%$),
but making a drastic and rather simplistic approximation that does not provide theoretical guarantees.

% \newpage
\section{Analysis and open problems \label{sec:analysis} } 
This section analyzes the independence-based algorithms surveyed, 
discussing their relative advantages and disadvantages,
and describes a series of open problems 
where future works may produce some advances in the area. 

\subsection{Analysis}

The independence-based algorithms for learning Markov networks 
are able to learn the independence structure efficiently,
having the important advantage of being sound (i.e., they guarantee to produce the true underlying distribution)
when data is a sample of a Markov network, tests are reliable, 
and the underlying distribution is strictly positive.
These algorithms perform a succession of statistical independence tests
to learn about the conditional independences present in the data,
and assume that those independences are satisfied in the underlying model.
The structure is learned by querying independences on data greedily, 
discarding all the structures that are inconsistent, until a single structure is left. 
An important source of errors in some of this algorithms 
is the cascade errors produced by erroneous statistical tests, that produces cumulatively errors. 
About their complexity, some of them can learn the structure by performing 
a polynomial number of tests in the number of variables $n$ of the domain.
This fact, together with the evidence that statistical tests may run in a time proportional
to the number of rows in the input dataset $D$,
result in some algorithms, e.g., GSMN, having a total execution time polynomial in $n$ and $D$.
When compared to score-based algorithms, 
they can learn the structure without the need of
an interleaved estimation of the numerical parameters of the model,
which is the main source of intractability of score-based algorithms for Markov networks.
The strength of independence-based algorithms is that they can learn correctly the structure under assumptions.
However, there is no equivalent theoretical guarantee for the correctness 
of the complete model resulting from learning the parameters for that structure. 

\begin{table}[ht]
\scriptsize
\center
	      \caption{Summary of independence-based Markov network approaches \label{tab:approaches}}
	      
		      \begin{tabular}{p{1.8cm}  p{3cm}  p{8cm}}
			    \multicolumn{3}{c}{~}\\
			    \hline
			    \\
			    \textbf{Name} & \textbf{Reference} & \textbf{Comments} \\ 
			    \\ \hline
			    \\

			    GSMN & \cite{Bromberg06} & \begin{minipage}{9cm}
					\begin{itemize}% \setlength{\itemsep}{-2mm} 
					\item Sound, under assumptions
					\item The first independence-based algorithm for Markov networks
					\item Use the local-to-global strategy 
					\item Performs a polynomial number of tests in the number \\ of variables of the domain $n$
					\item Quality depends on sample complexity of tests
					\end{itemize}
					\end{minipage} \\ 
			    \\ \hline
			    \\   
			    GSIMN &  \cite{Bromberg06}. & \begin{minipage}{9cm}
					\begin{itemize} % \setlength{\itemsep}{-2mm} 
					\item Sound, under assumptions
					\item Use the local-to-global strategy 
					\item Use Triangle theorem for reducing number of tests performed
					\item Useful when using large datasets, or distributed domains
					\item Savings up to $40\%$ in running times respect to GSMN
					\item Comparable quality respect to GSMN
					\end{itemize}
					\end{minipage} \\ 
			    \\ \hline
			    \\
			    PFMN & \cite{BROMBERG07} & \begin{minipage}{9cm}
					\begin{itemize} % \setlength{\itemsep}{-2mm} 
					\item Sound, under assumptions
					\item Does not use the local-to-global strategy 
					\item Designed for improving efficiency of GSIMN
					\item Use an approximate method for computing the \\ posterior $\Pr(G\mid D)$ using independence-tests
					\item Useful when using large datasets, or distributed domains
					\item Savings up to $90\%$ in running times respect to GSIMN
					\item Comparable quality respect to GSMN
					\end{itemize}
					\end{minipage} \\ 
			    \\ \hline
			    \\
			    DGSIMN & \cite{gandhi08} & \begin{minipage}{9cm}
					\begin{itemize} % \setlength{\itemsep}{-2mm} 
					\item Sound, under assumptions
					\item Use the local-to-global strategy 
					\item Designed for improving efficiency of GSIMN
					\item Use dynamic ordering for reducing number of tests performed
					\item Useful when using large datasets, or distributed domains
					\item Savings up to $85\%$ in running times respect to GSIMN
					\item Comparable quality respect to GSMN 
					\end{itemize}
					\end{minipage} \\ 
			    \\ \hline
			    \\
			    Argumentative test & \cite{BrombMarg09}& \begin{minipage}{9cm}
					\begin{itemize} % \setlength{\itemsep}{-2mm} 
					\item First approach for quality improvement
					\item Use argumentation to correct errors when tests are unreliable 
					\item Use an independence knowledge base. The inconsistencies \\ are used to detect errors in tests
					\item Designed for learning Bayesian and Markov networks. 
					\item Exact algorithm presented is exponential \\(improving accuracy up to $13\%$)
					\item Approximate algorithm proposed is simplistic, 
					and does not \\ provide theoretical guarantees \\ (improving accuracy up to $9\%$)
					\end{itemize}
					\end{minipage} \\ 
			    \\ \hline
			    \\

			\end{tabular}

\end{table}

The independence-based algorithms present in the literature for learning the structure of a Markov network
are GSMN, GSIMN, PFMN, DGSIMN. 
Related to structure learning is the argumentative independence test, 
an approach for improving the quality of conditional independences discovery.
Table~\ref{tab:approaches} shows a summary of the most important features of those approaches.
The GSMN algorithm is a direct extension of the GS algorithm for Markov networks structure learning,
which requires a polynomial number of tests in the number of variables of the domain $n$.
This algorithm is presented together with the GSIMN algorithm,
which improves the efficiency of GSMN by exploiting Pearl's independence axioms
to infer unknown independences from the independences observed so far,
avoiding the need of performing redundant statistical tests. This is important when 
datasets are large, or when datasets are present in distributed environments.
The results obtained for GSIMN show savings up to a $40\%$ in running times,
obtaining comparable qualities to GSMN.
The PFMN algorithm was designed for improving the efficiency of GSIMN.
This algorithm does not work in a local-to-global fashion.
Instead it uses a model for efficiently computing the approximate posterior probability of structures $\Pr(G \mid D)$. 
The results obtained by PFMN show improvements in running times up to $90\%$ with respect to GSIMN,
with equivalent quality of learned structures.
Similarly, the DGSIMN algorithm was designed for improving the efficiency of GSIMN, by enhancing the fixed ordering 
of variables in the Markov blanket learning subroutine
by a dynamic ordering mechanism. Experiments published for DGSIMN show improvements 
over the running times of GSIMN up to $85\%$, still maintaining the quality of GSMN. 

The most important problem of independence-based algorithms
for learning the structure of Markov networks is the problem of quality 
when statistical independence tests are not reliable. 
Such a problem is not tackled by either GSMN, GSIMN, DGSIMN or PFMN.
The only approach presented for improving the quality under uncertainty of tests outcomes 
is the argumentative independence test. Experimental results using this approach 
show significant improvements over the accuracy of the standard independence tests, 
but exact algorithms presented, while improving quality by $13\%$ have an exponential cost, 
and the approximate algorithm proposed   
make a drastic approximation that does not provide theoretical guarantees, 
still producing quality improvements, up to $9\%$.

In summary, the advantages of independence-based algorithms for learning Markov networks
are overshadowed by the low quality of learned structures when data is scarce, 
or equivalently, when the underlying network is highly connected.
This is why independence-based algorithms 
are not currently implemented in practice for learning Markov networks.
However, this approach presents important advantages that motivate further work in this area.
First, independence-based algorithms are sound (under assumptions) and efficient. 
Second, data availability is growing increasingly with time.
Third, there are several promising open problems (enumerated in the next section) 
whose solutions are expected to result in improvements in the quality of structures produced by this technology.

\subsection{Open Problems} \label{sec:openproblems}
Following the analysis of last section, this section discusses a series of open problems that 
remain in the area. 
All the listed problems focus on the quality and the efficiency 
of the independence-based approach for learning Markov networks.

\begin{description}
  \item 
	\begin{openproblem} \label{openproblem:ordering}
	  \textbf{Avoiding cascade errors}. 

	  Most independence-based algorithms surveyed (GSMN, GSIMN, DGSIMN) 
	  learn the Markov blanket of variables using
	  the GS algorithm. Learning the structure by using GS can be seen as a greedy search over the space of structures, 
	  where the outcomes of tests are used for discarding all those structures 
	  that are inconsistent with the independence indicated by the test. 
	  Therefore, an important source of errors in GS is the cascade errors produced by erroneous statistical tests,
	  that produces cumulatively errors. 

	  \begin{center}
	    \textbf{Is it possible to tackle the cascade effect, taking into account that tests are not always reliable?}
	  \end{center}
	\end{openproblem}

 \item 
	\begin{openproblem} \label{openproblem:qualitymeasure}
	  \textbf{Independence-based quality measures}. 
	  The PFMN algorithm uses the particle filter approach for optimizing the selection of tests to perform. 
	  It utilizes a generative model that computes approximately the posterior probability $\Pr(G\mid\data)$
	  of independence structures given the data.
	  Interestingly, this posterior probability can be efficiently computed, and can be used as a quality measure of candidate structures 
	  in an optimization method. This measure of structures quality has the advantage of avoiding cascade errors 
	  by assigning probabilities to structures.
	  This is an unexplored area for learning the structure of Markov networks.
	  \begin{center}
	    \textbf{Is it possible to adapt the structure posterior 
	    computation of PFMN into an efficient and sound score, by relaxing the approximation? 
	    Would the optimization of such score improve the quality of the structures learned?}
	  \end{center}
	\end{openproblem}

 \item 
	\begin{openproblem} \label{openproblem:speedup}
	  \textbf{Speeding up independence-based algorithms}. 
	  Learning the structure under the independence-based approach 
	  requires in some cases the execution of a massive amount 
	  of statistical independence tests on data. 
	  An intermediate step in the computation 
	  of independence tests is the construction of contingency tables from the dataset,
	  that record the frequency distribution of the variables involved in the test,
	  resulting in running times linear in the size of the dataset.

	  \begin{center}
	    \textbf{Can the contingency tables of some test be reused 
	    in the construction of contingency tables for other tests?
		    How can an independence-based algorithm use such a mechanism 
	    for minimizing the number of whole readings of the dataset?
		    Under what conditions would this mechanism generate gains in runtime complexity?
	    }
	  \end{center}
	\end{openproblem}

 \item \begin{openproblem} \label{openproblem:lgl}
	\textbf{Inconsistencies in local-to-global algorithms}. 
	Independence-based algorithms using the local-to-global strategy (e.g., GSMN and variants)
	decompose the problem of learning a complete independence structure 
	with $n$ variables into $n$ independent Markov blanket learning problems. 
	On a second step, these algorithms piece together all the learned Markov blankets into a global 
	structure using an ``OR rule''. Insufficient data may result in
	incorrect learning of Markov blankets, with conflicts in their decision on edge inclusion when,
	for two variables $X$ and $Y$, $X$ is found to be in the blanket of $Y$, but $Y$ is found not to be in the blanket of $X$. 
	In such cases, the ``OR rule'' always decides to add the edge, making mistakes when such edge
	does not exist. 
	  \begin{center}
	    \textbf{Is it possible to design more robust rules 
		    for solving inconsistencies between two learned Markov blankets?}
	  \end{center}
	\end{openproblem}

 \item 
	\begin{openproblem} \label{openproblem:comparison}
	\textbf{Comparing independence-based and score-based approaches}. 
	  There are several experimental comparisons currently lacking in the literature:
	  \begin{itemize}
	    \item There are no experimental results published comparing the sample complexity of both approaches.
	    \item There are no experimental results published comparing quality of structures learned by both approaches.
	    \item There are no experimental results published comparing quality of complete models: 
		  \textit{i)} those models learned by score-based approach 
		  (interleaving structure search and parameters estimation) versus 
		  \textit{ii)} models learned by independence-based approach 
		  (learning the structure and then fitting the parameters only once for such structure).
	  \end{itemize}
	  \begin{center}
	    \textbf{Are the independence-based algorithms valid as a practical alternative to score-based algorithms for learning the structure,
		    and for learning the complete model?
		    }
	  \end{center}

	\end{openproblem}

  \item 
	\begin{openproblem} \label{openproblem:adaptingBN}
	\textbf{Adapting recent Bayesian network ideas to Markov networks}.       
	  The first independence-based algorithm proposed is GSMN, an adaptation to Markov networks of the GS algorithm.
	  In the literature there are several recently proposed ideas 
	  for improving the efficiency, quality and sample complexity of GS,
	  as those discussed by the authors of 
	  IAMB. MMPC/MB, HITON-PC/MB, Fast-IAMB, PCMB and IPC-MB algorithms 
	  (see Section~\ref{sec:independenBased}, for more details).
	  However, all these interesting ideas are 
	  originally developed and tested for learning 
	  the structure of Bayesian networks.
	  \begin{center}
	    \textbf{Can this research be adapted to the Markov networks structure learning problem,
	    to generate some improvements in the area?
	    }
	  \end{center}
	\end{openproblem}

 \item 
	\begin{openproblem} \label{openproblem:ikb}
	  \textbf{Independence knowledge bases}.  
	  The argumentative independence test
	  improves the accuracy of tests significantly when data is scarce. 
	  However, the exact algorithm proposed by this approach
	  runs in exponential time, because Pearl's axioms are in first-order logics,
	  and knowledge bases in argumentation are propositional (as detailed in Section~\ref{sec:argumentation}). 
	  The approximate solution presented is polynomial in 
	  running time, still improving the quality,
	  but making a drastic and rather simplistic approximation that does not provide theoretical guarantees.
	   \begin{center}
	    \textbf{Can the Pearl's axioms be exploited in a more efficient manner, 
		  through a better approximation,
		  by an alternative formalism for reasoning under inconsistencies?}
	  \end{center}
	\end{openproblem}

  \item 
	\begin{openproblem} \label{openproblem:independentindependences}
	\textbf{Relating independence assertions}.  
	  Statistical tests are procedures that run independently to each other,
	  and they are used as a black box by independence-based algorithms. 
	  Each test responds to a conditional independence query only using the input dataset.
	  Thus, an implicit assumption made by all the independence-based algorithms
	  is that all the independences queried by the algorithm are mutually independent to each other given the dataset.
	  This assumption is only true when data is sufficiently large for the test to determine
	  the true underlying independence, because in this case, information from other tests is irrelevant.
	  However, when data is not sufficient for correctly determining the independence,
	  tests may become dependent given the data, i.e., information from other tests may be useful 
	  for determining the value of a test, and avoiding errors. 
	  An example shown in the literature for correcting errors when data is insufficient 
	  is the argumentative independence test, that relates statistical tests through
	  the Pearl's axioms, as additional information for improving the quality of tests when data is not sufficient.
	  \begin{center}
	    \textbf{Besides Pearl's axioms, are there other dependence relations governing independence assertions? 
		As in the case of Pearl's axioms, can these relations be used as additional information 
		for improving the quality of independence-based algorithms?}
	  \end{center}
	\end{openproblem}

\end{description}

\section{Conclusions}
The present work discussed the most relevant technical aspects in the problem of learning 
the Markov network structure from data,
stressing on independence-based algorithms.
Summarizing the analysis of such technology, the advantages of independence-based algorithms for learning Markov networks
are overshadowed by the low quality of learned structures when data is scarce, 
or equivalently, when the underlying network is highly connected.
However, this approach presents important advantages that motivate further work in this area.
First, independence-based algorithms are sound under assumptions, and efficient. 
Second, data availability is growing increasingly with time.
Therefore, it is expected that the solutions of the open problems posed in this work 
result in improvements in the quality of structures produced by this technology.

\newpage
\bibliographystyle{spbasic}
\bibliography{SURVEY-IB4MNs2012-SCHLUTER-BROMBERG}

\begin{thebibliography}{63}
\providecommand{\natexlab}[1]{#1}
\providecommand{\url}[1]{{#1}}
\providecommand{\urlprefix}{URL }
\expandafter\ifx\csname urlstyle\endcsname\relax
  \providecommand{\doi}[1]{DOI~\discretionary{}{}{}#1}\else
  \providecommand{\doi}{DOI~\discretionary{}{}{}\begingroup
  \urlstyle{rm}\Url}\fi
\providecommand{\eprint}[2][]{\url{#2}}

\bibitem[{Agresti(2002)}]{AGRESTI02}
Agresti A (2002) {Categorical Data Analysis}, 2nd edn. Wiley

\bibitem[{Alden(2007)}]{Alden07}
Alden M (2007) {MARLEDA: Effective Distribution Estimation Through Markov
  Random Fields}. PhD thesis, Dept of CS, University of Texas Austin

\bibitem[{Aliferis et~al(2003)Aliferis, Tsamardinos, and
  Statnikov}]{Aliferis2003:HITON}
Aliferis C, Tsamardinos I, Statnikov A (2003) {HITON}, a novel {Markov} blanket
  algorithm for optimal variable selection. AMIA Fall

\bibitem[{Aliferis et~al(2010{\natexlab{a}})Aliferis, Statnikov, Tsamardinos,
  Mani, and Koutsoukos}]{Aliferis2010}
Aliferis C, Statnikov A, Tsamardinos I, Mani S, Koutsoukos X
  (2010{\natexlab{a}}) {Local Causal and Markov Blanket Induction for Causal
  Discovery and Feature Selection for Classification Part I: Algorithms and
  Empirical Evaluation}. JMLR 11:171--234

\bibitem[{Aliferis et~al(2010{\natexlab{b}})Aliferis, Statnikov, Tsamardinos,
  Mani, and Koutsoukos}]{Aliferis2010b}
Aliferis C, Statnikov A, Tsamardinos I, Mani S, Koutsoukos X
  (2010{\natexlab{b}}) {Local Causal and Markov Blanket Induction for Causal
  Discovery and Feature Selection for Classification Part II: Analysis and
  Extensions}. JMLR 11:235--284

\bibitem[{Amgoud and Cayrol(2002)}]{Amgoud2002:arg}
Amgoud L, Cayrol C (2002) {A Reasoning Model Based on the Production of
  Acceptable Arguments}. Annals of Mathematics and Artificial Intelligence
  34:197--215

\bibitem[{Anguelov et~al(2005)Anguelov, Taskar, Chatalbashev, Koller, Gupta,
  Heitz, and Ng}]{ANGUELOVTASKAR05}
Anguelov D, Taskar B, Chatalbashev V, Koller D, Gupta D, Heitz G, Ng A (2005)
  {Discriminative Learning of {Markov} Random Fields for Segmentation of {3D}
  Range Data}. Proceedings of the CVPR

\bibitem[{Barahona(1982)}]{BARAHONA82}
Barahona F (1982) On the computational complexity of {Ising} spin glass models.
  Journal of Physics A: Mathematical and General 15(10):3241--3253

\bibitem[{Besag(1977)}]{besag77}
Besag J (1977) Efficiency of pseudolikelihood estimation for simple {G}aussian
  fields. Biometrica 64:616--618

\bibitem[{Besag et~al(1991)Besag, York, and Mollie}]{BESAG91}
Besag J, York J, Mollie A (1991) {Bayesian} image restoration with two
  applications in spatial statistics. Annals of the Inst of Stat Math 43:1--59

\bibitem[{Bromberg(2007)}]{Bromberg08:thesis}
Bromberg F (2007) {Markov network structure discovery using independence
  tests}. PhD thesis, Dept of CS, Iowa State University

\bibitem[{Bromberg and Margaritis(2007)}]{BROMBERG07}
Bromberg F, Margaritis D (2007) Efficient and robust independence-based
  {Markov} network structure discovery. In: Proceedings of IJCAI

\bibitem[{Bromberg and Margaritis(2009)}]{BrombMarg09}
Bromberg F, Margaritis D (2009) {Improving the Reliability of Causal Discovery
  from Small Data Sets using Argumentation}. JMLR 10:301--340

\bibitem[{Bromberg et~al(2006)Bromberg, Margaritis, and Honavar}]{Bromberg06}
Bromberg F, Margaritis D, Honavar V (2006) {Efficient markov network structure
  discovery using independence tests}. In: In Proc SIAM Data Mining, p~06

\bibitem[{Bromberg et~al(2009)Bromberg, Margaritis, and
  V.}]{brombergmargaritis09b}
Bromberg F, Margaritis D, V H (2009) {Efficient Markov Network Structure
  Discovery Using Independence Tests}. JAIR 35:449--485

\bibitem[{Cai et~al(2007)Cai, Bu, Chen, and Qiu}]{cai07}
Cai Kk, Bu Jj, Chen C, Qiu G (2007) A novel dependency language model for
  information retrieval. Journal of Zhejiang University - Science A 8:871--882,
  10.1631/jzus.2007.A0871

\bibitem[{Cochran(1954)}]{COCHRAN54}
Cochran WG (1954) Some methods of strengthening the common ${\chi}$ tests.
  Biometrics p 10:417–451

\bibitem[{Cooper(1990)}]{Cooper1990}
Cooper GF (1990) {The computational complexity of probabilistic inference using
  bayesian belief networks}. Artificial Intelligence 42(2-3):393 -- 405,
  \doi{DOI: 10.1016/0004-3702(90)90060-D}

\bibitem[{Cover and Thomas(1991)}]{cover&tomas91}
Cover TM, Thomas JA (1991) {Elements of information theory}.
  Wiley-Interscience, New York, NY, USA

\bibitem[{Cressie(1992)}]{cressie92}
Cressie N (1992) Statistics for spatial data. Terra Nova 4(5):613--617,
  \doi{10.1111/j.1365-3121.1992.tb00605.x}

\bibitem[{Davis and Domingos(2010)}]{DavisAndDomingos2010:BottomUp}
Davis J, Domingos P (2010) {Bottom-Up Learning of Markov Network Structure}.
  In: ICML, pp 271--278

\bibitem[{Della~Pietra et~al(1997)Della~Pietra, Della~Pietra, and
  Lafferty}]{PietraPL97}
Della~Pietra S, Della~Pietra VJ, Lafferty JD (1997) {Inducing Features of
  Random Fields}. IEEE Trans PAMI 19(4):380--393

\bibitem[{Friedman et~al(2000)Friedman, Linial, Nachman, and
  Pe'er}]{friedman00}
Friedman N, Linial M, Nachman I, Pe'er D (2000) {Using {Bayesian} Networks to
  Analyze Expression Data}. Computational Biology 7:601--620

\bibitem[{Fu and Desmarais(2008)}]{Fu2008:IPCMB}
Fu S, Desmarais MC (2008) {Fast Markov blanket discovery algorithm via local
  learning within single pass}. In: {Proceedings of the Canadian Society for
  computational studies of intelligence, 21st conference on Advances in
  artificial intelligence}, Springer-Verlag, Berlin, Heidelberg, Canadian
  AI'08, pp 96--107

\bibitem[{Fu and Desmarais(2010)}]{fu2010:reviewMB}
Fu S, Desmarais MC (2010) {Markov Blanket based Feature Selection : A Review of
  Past Decade}. Proceedings of the World Congress on Engineering 2010
  I:321--328

\bibitem[{Ganapathi et~al(2008)Ganapathi, Vickrey, Duchi, and
  Koller}]{ganapathi2008}
Ganapathi V, Vickrey D, Duchi J, Koller D (2008) {Constrained Approximate
  Maximum Entropy Learning of Markov Random Fields}. In: Uncertainty in
  Artificial Intelligence, pp 196--203

\bibitem[{Gandhi et~al(2008)Gandhi, Bromberg, and Margaritis}]{gandhi08}
Gandhi P, Bromberg F, Margaritis D (2008) {Learning Markov Network Structure
  using Few Independence Tests}. In: {SIAM International Conference on Data
  Mining}, pp 680--691

\bibitem[{Heckerman et~al(1995)Heckerman, Geiger, and Chickering}]{Heckerman95}
Heckerman D, Geiger D, Chickering DM (1995) {Learning {Bayesian} Networks: The
  Combination of Knowledge and Statistical Data}. Machine Learning

\bibitem[{H{\"o}fling and Tibshirani(2009)}]{Hofling09}
H{\"o}fling H, Tibshirani R (2009) {Estimation of Sparse Binary Pairwise Markov
  Networks using Pseudo-likelihoods}. Journal of Machine Learning Research
  10:883--906

\bibitem[{Hyv{\"a}rinen and Dayan(2005)}]{Hyvarinen05}
Hyv{\"a}rinen A, Dayan P (2005) {Estimation of non-normalized statistical
  models by score matching}. Journal of Machine Learning Research 6:695--709

\bibitem[{Karyotis(2010)}]{Karyotis2010}
Karyotis V (2010) Markov random fields for malware propagation: the case of
  chain networks. Comm Letters 14:875--877

\bibitem[{Koller and Friedman(2009)}]{koller09}
Koller D, Friedman N (2009) {Probabilistic Graphical Models: Principles and
  Techniques}. MIT Press

\bibitem[{Koller and Sahami(1996)}]{KollerSahami1996:KS}
Koller D, Sahami M (1996) {Toward Optimal Feature Selection}. Morgan Kaufmann,
  pp 284--292

\bibitem[{Lam and Bacchus(1994)}]{LamBacchus94}
Lam W, Bacchus F (1994) Learning {Bayesian} belief networks: an approach based
  on the {MDL} principle. Computational Intelligence 10:269--293

\bibitem[{Larra{\~{n}}aga and Lozano(2002)}]{larranagaANDlozano2002}
Larra{\~{n}}aga P, Lozano JA (2002) {E}stimation of {D}istribution
  {A}lgorithms. {A} {N}ew {T}ool for {E}volutionary {C}omputation. Kluwer Pubs

\bibitem[{Lauritzen(1996)}]{LAURITZEN96}
Lauritzen SL (1996) Graphical Models. Oxford University Press

\bibitem[{Lee et~al(2006)Lee, Ganapathi, and Koller}]{Lee+al:NIPS06}
Lee SI, Ganapathi V, Koller D (2006) Efficient structure learning of {M}arkov
  networks using {L1}-regularization. In: NIPS

\bibitem[{Li(2001)}]{Li2006}
Li SZ (2001) {Markov random field modeling in image analysis}. Springer-Verlag
  New York, Inc., Secaucus, NJ, USA

\bibitem[{Margaritis(2005)}]{MARGARITIS05}
Margaritis D (2005) {Distribution-Free Learning of Bayesian Network Structure
  in Continuous Domains}. In: Proceedings of AAAI

\bibitem[{Margaritis and Bromberg(2009)}]{margaritisBromberg09}
Margaritis D, Bromberg F (2009) {Efficient Markov Network Discovery Using
  Particle Filter}. Comp Intel 25(4):367--394

\bibitem[{Margaritis and Thrun(2000)}]{MARGARITIS00}
Margaritis D, Thrun S (2000) {Bayesian} network induction via local
  neighborhoods. In: Proceedings of NIPS

\bibitem[{McCallum(2003)}]{MCCALLUM03}
McCallum A (2003) Efficiently inducing features of conditional random fields.
  In: Proceedings of Uncertainty in Artificial Intelligence (UAI)

\bibitem[{Metzler and Croft(2005)}]{Metzler05amarkov}
Metzler D, Croft WB (2005) A markov random field model for term dependencies.
  In: Proceedings of the 28th annual international ACM SIGIR conference on
  Research and development in information retrieval, ACM, New York, NY, USA,
  SIGIR '05, pp 472--479

\bibitem[{Minka(2001)}]{Minka_2001}
Minka T (2001) Algorithms for maximum-likelihood logistic regression. Tech.
  rep., Dept of Statistics, Carnegie Mellon University

\bibitem[{Minka(2004)}]{minka2004}
Minka T (2004) {Power EP}. Tech. Rep. MSR-TR-2004-149, Microsoft Research,
  Cambridge

\bibitem[{Mooij(2010)}]{Mooij08libdai}
Mooij JM (2010) {lib{DAI}: A Free and Open Source C++ Library for Discrete
  Approximate Inference in Graphical Models}. J Mach Learn Res 11:2169--2173

\bibitem[{Pearl(1988)}]{pearl88}
Pearl J (1988) {Probabilistic Reasoning in Intelligent Systems: Networks of
  Plausible Inference}. Morgan Kaufmann Publishers, Inc.

\bibitem[{Pearl and Paz(1985)}]{PearlPaz1985:graphoids}
Pearl J, Paz A (1985) {GRAPHOIDS : A graph based logic for reasonning about
  relevance relations}. Tech. Rep. 850038 (R-53-L), Cognitive Systems
  Laboratory, University of California, Los Angeles

\bibitem[{Pe{\~{n}}a et~al(2007)Pe{\~{n}}a, Nilsson, {Bj{\"o}rkegren}, and
  Tegn{\'e}r}]{penia2007:PCMB}
Pe{\~{n}}a JM, Nilsson R, {Bj{\"o}rkegren} J, Tegn{\'e}r J (2007) {Towards
  scalable and data efficient learning of Markov boundaries.} Int J Approx
  Reasoning pp 211--232

\bibitem[{Ravikumar et~al(2010)Ravikumar, Wainwright, and
  Lafferty}]{ravikumar2010:l1}
Ravikumar P, Wainwright MJ, Lafferty JD (2010) {High-dimensional Ising model
  selection using L1-regularized logistic regression}. Annals of Statistics
  38:1287--1319, \doi{10.1214/09-AOS691}

\bibitem[{Schmidt et~al(2008)Schmidt, Murphy, Fung, and Rosales}]{schmidts08}
Schmidt M, Murphy K, Fung G, Rosales R (2008) Structure learning in random
  fields for heart motion abnormality detection. In: Computer Vision and
  Pattern Recognition, 2008. CVPR 2008. IEEE Conference on, pp 1 --8,
  \doi{10.1109/CVPR.2008.4587367}

\bibitem[{Shakya and Santana(2008)}]{moapaper}
Shakya S, Santana R (2008) A markovianity based optimization algorithm. Tech.
  rep., Basque Country U.

\bibitem[{Shekhar et~al(2004)Shekhar, Zhang, Huang, and Vatsavai}]{shekhar04}
Shekhar S, Zhang P, Huang Y, Vatsavai RR (2004) {Trends in Spatial Data
  Mining}. In: Kargupta H, Joshi A, Sivakumar K, Yesha Y (eds) {Trends in
  Spatial Data Mining}, AAAI Press / The MIT Press, chap~19, pp 357--379

\bibitem[{Spirtes et~al(2000)Spirtes, Glymour, and Scheines}]{Spirtes00}
Spirtes P, Glymour C, Scheines R (2000) {Causation, Prediction, and Search}.
  Adaptive Computation and Machine Learning Series, MIT Press

\bibitem[{Tsamardinos et~al(2003)Tsamardinos, Aliferis, and
  Statnikov}]{tsamardinos03:IAMB}
Tsamardinos I, Aliferis CF, Statnikov A (2003) Algorithms for large scale
  {Markov} blanket discovery. In: FLAIRS

\bibitem[{Tsamardinos et~al(2006)Tsamardinos, Brown, and
  Aliferis}]{Tsamardinos2006:MMHC}
Tsamardinos I, Brown L, Aliferis CF (2006) The max-min hill-climbing {Bayesian}
  network structure learning algorithm. Machine Learning 65:31--78

\bibitem[{Vishwanathan et~al(2006)Vishwanathan, Schraudolph, Schmidt, and
  Murphy}]{Vishwanathan2006}
Vishwanathan SVN, Schraudolph NN, Schmidt MW, Murphy KP (2006) {Accelerated
  training of conditional random fields with stochastic gradient methods}. In:
  Proceedings of the 23rd international conference on Machine learning, ACM,
  New York, NY, USA, ICML '06, pp 969--976

\bibitem[{Wainwright and Jordan(2008)}]{Wainwright2008}
Wainwright MJ, Jordan MI (2008) {Graphical Models, Exponential Families, and
  Variational Inference}. Found Trends Mach Learn 1:1--305,
  \doi{10.1561/2200000001}

\bibitem[{Wainwright et~al(2003)Wainwright, Jaakkola, and
  Willsky}]{Wainwright03tree-reweightedbelief}
Wainwright MJ, Jaakkola TS, Willsky AS (2003) Tree-reweighted belief
  propagation algorithms and approximate {ML} estimation by pseudo-moment
  matching. In: In AISTATS

\bibitem[{Winn and Bishop(2005)}]{Winn2005:VMP}
Winn J, Bishop CM (2005) {Variational Message Passing}. J Mach Learn Res
  6:661--694

\bibitem[{Yaramakala and Margaritis(2005)}]{Yaramakala2005:fastIAMB}
Yaramakala S, Margaritis D (2005) {Speculative Markov blanket discovery for
  optimal feature selection}. In: {Data Mining, Fifth IEEE International
  Conference on}, p 4 pp., \doi{10.1109/ICDM.2005.134}

\bibitem[{Yedidia et~al(2005)Yedidia, Freeman, and Weiss}]{yedidia2005}
Yedidia J, Freeman W, Weiss Y (2005) Constructing free-energy approximations
  and generalized belief propagation algorithms. Information Theory, IEEE
  Transactions on 51(7):2282 -- 2312, \doi{10.1109/TIT.2005.850085}

\bibitem[{Yedidia et~al(2004)Yedidia, Freeman, and
  Weiss}]{Yedidia04constructingfree}
Yedidia JS, Freeman WT, Weiss Y (2004) {C}onstructing {F}ree {E}nergy
  {A}pproximations and {G}eneralized {B}elief {P}ropagation {A}lgorithms. IEEE
  Transactions on Information Theory 51:2282--2312

\end{thebibliography}

\end{document}